\documentclass{article}
\usepackage{ijcai20}

\usepackage{times}

\usepackage{url}
\usepackage[hidelinks]{hyperref}
\usepackage[utf8]{inputenc}
\usepackage[small]{caption}
\usepackage{graphicx}
\usepackage{amsmath}
\usepackage{booktabs}
\title{Formal Query Building with Query Structure Prediction for Complex Question Answering over Knowledge Base}

\author{
Yongrui Chen\and
Huiying Li\footnote{Contact Author}\and
Yuncheng Hua\And
Guilin Qi\\
\affiliations
School of Computer Science and Engineering, Southeast University, Nanjing, China\\
\emails
\{220171664, huiyingli, devinhua, gqi\}@seu.edu.cn,
}

\begin{document}

\maketitle

\begin{abstract}
Formal query building is an important part of complex question answering over knowledge bases. It aims to build correct executable queries for questions. Recent methods try to rank candidate queries generated by a state-transition strategy. However, this candidate generation strategy ignores the structure of queries, resulting in a considerable number of noisy queries. In this paper, we propose a new formal query building approach that consists of two stages. In the first stage, we predict the query structure of the question and leverage the structure to constrain the generation of the candidate queries. We propose a novel graph generation framework to handle the structure prediction task and design an encoder-decoder model to predict the argument of the predetermined operation in each generative step. In the second stage, we follow the previous methods to rank the candidate queries. The experimental results show that our formal query building approach outperforms existing methods on complex questions while staying competitive on simple questions.
\end{abstract}

\section{Introduction}
Knowledge Base Question Answering (KBQA) is an active research area where the goal is to provide crisp answers to natural language questions. An important direction in  KBQA is answering via semantic parsing ~\cite{DBLP:conf/coling/BaoDYZZ16,DBLP:conf/emnlp/LuoLLZ18,DBLP:conf/acl/YihCHG15} to natural language questions, transforming the corresponding semantic components, including entities, relations and various constraints, into formal queries (e.g., SPARQL) and then executing queries over the knowledge base (KB) to retrieve answers. In general, a question can be regarded as correctly answered if its correct query has been built. For a complex question, whose query consists of multiple entities, relations, and constraints, how to select the correct semantic components of the query and combine them in an appropriate way is still a problem to be solved.
	
	\begin{figure}
		\centerline{\includegraphics[width=0.5\textwidth]{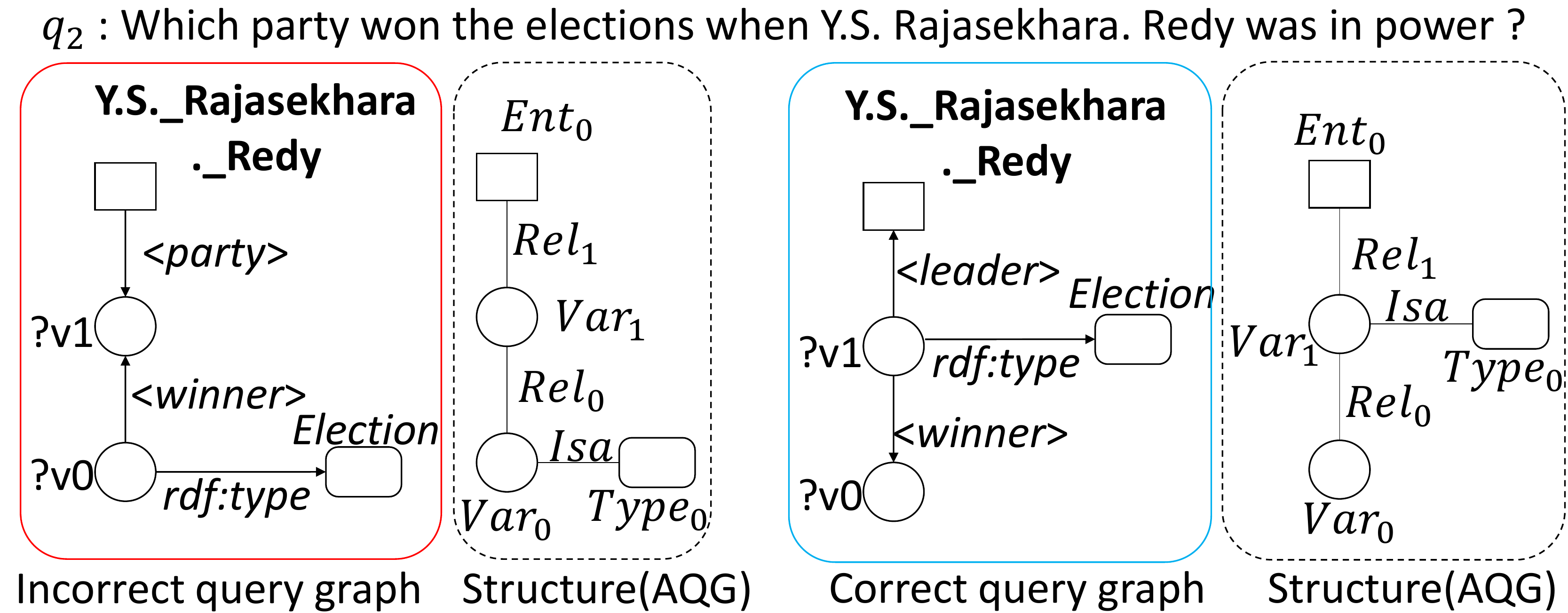}}
		\caption{ One example to prove the importance of query structures. For each query graph, ?v0 is the placeholder of the answer and ?v1 (if any) is the hidden variable.} \label{one_question}
	\end{figure}
	
	In recent research, most query building approaches~\cite{DBLP:conf/emnlp/LuoLLZ18,DBLP:conf/semweb/MaheshwariTLCF019} are based on query ranking, i.e., scoring candidate queries by similarities with the question and outputting the top-scored one. Such ranking-based approaches leverage a \textit{query graph} to represent formal queries and generate candidate query graphs by a state-transition strategy \cite{DBLP:conf/acl/YihCHG15,DBLP:conf/emnlp/LuoLLZ18,DBLP:conf/semweb/MaheshwariTLCF019}. This strategy first assumes that the length of the path between the topic entity\footnote{``topic entity" is proposed in \cite{DBLP:conf/acl/YihCHG15}, which refers to the root of the query graph(tree).} and the answer does not exceed two hops, and then generates the candidate queries by enumerating all possible structures meeting this assumption. Although the strategy can cover almost all correct query graphs in existing KBQA datasets, it has a significant drawback: a considerable number of noisy query graphs are produced. These noisy queries have incorrect structures but their components have high similarity to the question. In fact, existing query ranking models often make erroneous judgments when dealing with them. For example, Figure \ref{one_question} shows one question that was answered incorrectly by the states-of-the-arts ranking-based approach\footnote{We reimplement the approach proposed in \cite{DBLP:conf/emnlp/LuoLLZ18}.}. Below the question are two candidate query graphs generated by the above strategy. The model does not identify the structure (dotted box) of the correct query (blue box) so that it selects the noisy query (red box). 
	
	To avoid producing noisy queries, we propose abstract query graphs (AQG) to describe query structures and propose a new formal query building approach with query structure prediction. Specifically, our approach consists of two stages. In the first stage, we first leverage a neural network-based generative model to generate an AQG according to the question. The AQG is regarded as a predicted structure of the query graph. Then, we utilize the AQG as a constraint to generate the candidate query graphs whose structures match it. In the second stage, we perform candidate query ranking by an existing model, in order to evaluate the improvement brought by the AQG constraint. Following the previous approaches, our work focuses on handling the query graph of the tree structure. We perform comprehensive experiments on multiple QA datasets, and our proposed method consistently outperforms previous approaches on complex questions and produces competitive results on the dataset mainly made up of simple questions.

\section{Preliminaries} \label{preliminaries}
\subsection{Query Graph}
	The \textit{knowledge base} (KB) can be regarded as a collection of subject-predicate-object triples $\left \langle s, p, o\right \rangle$, where $s$ is an entity, $p$ is a relation and $o$ can be either a literal or an entity. 
	
	The \textit{query graph} is defined in \cite{DBLP:conf/acl/YihCHG15}, which is a structured representation of natural language questions. In theory, it is a directed graph. However, in the existing datasets, it is often a tree structure with $n$ vertices and $n-1$ edges. Therefore, we focus on dealing with the queries of the tree structure in this paper, and the process to general graph structures is left for future work. The vertices of query graphs include KB entities, KB types, etc., and the edges consist of KB relations and some built-in properties. Table \ref{query_graph} shows a detail classification of the vertices (middle) and edges (bottom).
	
    	\begin{table}
		\begin{center}
			\scalebox{0.8}{
			\begin{tabular}{ccc}
				\hline
				\rule{0pt}{12pt}
				Class & Instances & Symbol
				\\
				\hline
				\\[-6pt]
				Entity & \textbf{Y.S.\_Rajasekhara.\_Redy}, $\cdots$ & rectangle \\
				Type & \textit{Election}, \textit{Actor}, \textit{President}, $\cdots$ & rounded rectangle \\
				Number & \textbf{1}, \textbf{2008}, $\cdots$ & diamond \\
				Variable & ?v0, ?v1, $\cdots$ & circle \\
				\hline
				\\[-6pt]
				Relation & $\langle$\textit{party}$\rangle$, $\langle$\textit{winner}$\rangle$, $\cdots$ & - \\
				Order & \textit{min\_at\_n}, \textit{max\_at\_n} & - \\
				Comparison & $<$, $>$, $=$ & - \\
				Count & count & - \\
				Isa & \textit{rdf:type} & - \\
				\hline
				\\[-6pt]
			\end{tabular}}
			{\caption{Classes of vertices and edges in query graphs}\label{query_graph}}
		\end{center}
	\end{table}
	
\subsection{Abstract Query Graph}
	Following the work given in \cite{DBLP:conf/emnlp/Hu0Z18,DBLP:conf/emnlp/DingHXQ19}, we propose an \textit{abstract query graph} to describe the query structure. An abstract query graph (AQG) is a tree, which consists of a combination of $n$ labeled vertices $v \in $ \{``Ent", ``Type", ``Num", ``Var"\}, connected with $n-1$ labeled, undirected edges $e \in $ \{``Rel", ``Ord", ``Cmp", ``Cnt", ``Isa"\}. Here, all the labels in AQG correspond to the vertex classes and edge classes in query graphs (see Table \ref{query_graph}). Intuitively, AQG reflects the topology and the component classes of query graphs and each query graph corresponds to a unique AQG (Figure \ref{one_question}). Therefore, if the correct AQG is predicted, it can be used as a constraint to avoid generating the noisy queries with wrong structures.  
\subsection{Grounding}
	The operation of utilizing AQG to generate candidate queries is called \textit{grounding}. In our work, we perform grounding on an AQG $g$ by a two-step method: first, we obtain an intermediate graph by replacing all the vertices (not variables) and built-in property edges of $g$ with the candidate instances. The intermediate graph is denoted by $g^*$. Then, we execute the queries\footnote{The queries are generated from $g^*$ and all the possible directions of ``Rel" edges are taken into account. That is, if there are two ``Rel" edges in the AQG, the number of the direction combinations is 4.} (SPARQL) corresponding to $g^*$ against the KB to retrieve all instance relations corresponding to each "Rel" edges of $g^*$. After replacing the ``Rel" edges of $g^*$ with the retrieved instance relations, the query graph is generated. Concretely, for each invariable vertex, its candidate instances are obtained from the existing linking results, including entities, type, number (time and ordinal). For each edge, if its class label is ``Rel", its candidate instances retrieved by executing $g^*$. Otherwise, it is a built-in property and its candidate instances are obtained by referring to Table \ref{query_graph}. After all the combinations of the candidate instances are tried to replace vertices and edges, and their legitimacy is verified by the KB, grounding is over and the candidate query graphs corresponding to $g$ are finally generated.
	
\subsection{Graph Transformer}
	Since query structure is a tree or even a graph, traditional neural networks cannot effectively capture its structural information. In this paper, we introduce graph transformer \cite{DBLP:conf/naacl/Koncel-Kedziorski19}, which is a graph neural network, to learn the vector representation of AQG. Its input is a multi-labeled graph (e.g., AQG), which consists of three components: a sequence of vertices, a sequence of edges, and an adjacency matrix. Here, both sequences are unordered and each vertex or edge is represented by its corresponding class label vector, which is randomly initialized. For each vertex or edge in the graph, graph transformer exploits the multi-head attention mechanism \cite{DBLP:conf/nips/VaswaniSPUJGKP17} to aggregate the information of its neighbors (within several hops) to update its vector representation. Finally, the output includes three parts: a sequence of vertex vectors, a sequence of edge vectors, and a global vector of the entire graph, which is obtained by the weighted sum of all the vertices and edges.

\section{Approach}
	
	\subsection{Process Overview}
	For an input question $q$, we first assume that the linking results of entities, type, and number in $q$ have been obtained by preprocessing, denoted by $R_l$. Then, the query graph $g_{q}$ corresponding to $q$ is built by the following two stages:
	
	\begin{itemize}
		\item[(a)] \textbf{Generating candidate query graphs.} First, an AQG generator translates $q$ into an AQG, denoted by $g$. Then, according to $R_l$, a query graph candidate set $C_g$ is generated by performing grounding on $g$.
		
		\item[(b)] \textbf{Ranking candidate query graphs.} An existing query ranking model is employed as a black box to score each query graph in $C_g$ and the top-scored one is selected as $g_{q}$.
	\end{itemize}
	
	Our approach aims to explore how to predict query structures (AQG generation) and whether they can improve existing query ranking approaches.  Consequently, in order for the experimental comparison, we keep the query ranking model consistent with \cite{DBLP:conf/emnlp/LuoLLZ18}. To more intuitively show the value of AQG and shorten the training time of the ranking model, we only employ their basic version model which removes the dependency representation of the question and id representation of the path (refer to \cite{DBLP:conf/emnlp/LuoLLZ18} for details). In the rest of this section, we will detail the process of AQG generation.
	
	\subsection{Abstract Query Graph Generation}
    We first introduce the proposed generative framework, and then detail the neural network-based AQG generator which follows this framework.
	\label{section_geration}
	\subsubsection{Generative Framework}
	An AQG is a tree, which can be regarded as a specific case of undirected graphs, denoted by $g = (V, E)$. Here, $V$ and $E$ are the sets of labeled vertices and labeled edges, respectively. In our framework, the generative process of AQG can be described by a sequence of graphs $G = \left\{ g^0, g^1,..., g^L \right\}$, where $g^0$ is an empty graph and $g^L$ is the completed AQG. For each time step $t \ge 1$, $g^t = f(g^{t-1}, *a^t)$, where $f$ denotes the operator and $*a^t$ denotes several arguments according to $f$. In this paper, we define the following three types of graph-level operators: 
	
	\begin{itemize}
		\item[(a)] \textit{addVertex.} For a graph $g^t = (V, E)$, \textit{addVertex}($g^t$, $c_v$) represents adding a fresh vertex of label $c_v$ into $g^t$. The result is a graph $g^{t+1} = (V \cup \left\{ v_{add} \right\}, E)$, where $v_{add}$ denotes the added vertex.
		
		\item[(b)] \textit{selectVertex.} For the graph $g^{t+1} = (V \cup \left\{ v_{add} \right\}, E)$ after \textit{addVertex},  \textit{selectVertex}($g^{t+1}$, $v_{add}$, $v_{slc}$) means selecting a vertex $v_{slc} \in V$ that will be connected to the fresh vertex $v_{add}$. The results is a graph $g^{t+2} = (V \cup \left\{ v_{add} \right\}, E)$\footnote{Note that the structures of $g^{t+1}$ and $g^{t+2}$ are identical. Here, utilizing the two different identifiers aims to distinguish the time steps they belong to.}.
		
		\item[(c)] \textit{addEdge.} For the graph $g^{t+2} = (V \cup \left\{ v_{add} \right\}, E)$ after \textit{selectVertex}, \textit{addEdge}($g^{t+2}$, $v_{add}$, $v_{slc}$, $c_e$) means adding a fresh edge of label $c_e$ into $g^{t+2}$ to connect $v_{add}$ with $v_{slc}$. The result is a graph $g^{t+3} = (V \cup \left\{ v_{add} \right\}, E \cup \left\{ e_{add} \right\} )$, where $e_{add} = (v_{slc}, c_e, v_{add})$ is the added edge.
		
	\end{itemize}
	
	Successively performing the operations \textit{addVertex}, \textit{selectVertex}, and \textit{addEdge} on a graph is called an iteration, which adds a new triple $\langle v_{slc}, c_e, v_{add} \rangle$ into the graph. In the proposed framework, we stipulate that the first operation is always \textit{addVertex}, which converts the initial empty graph to a graph containing only one isolated vertex. Then, the AQG is generated by multiple iterations. Although the operator $f$ is predetermined in each time step, different arguments $*a^t$ result in generating different AQG. Here, the arguments $v_{add}$ in \textit{selectVertex}, $v_{add}$ and $v_{slc}$ in \textit{addEdge} can be obtained by the previous step, consequently, only the last argument of each operation need be determined, i.e., $c_v$ in \textit{addVertex}, $v_{slc}$ in \textit{selectVertex} and $c_e$ in \textit{addEdge}. 
	
	There are two reasons for predicting query structures by a grammar (graph-level operations) -based generation process rather than predefined structure templates: (1) First, compared to the templates, the graph-level operations are more underlying and therefore more flexible. (2) Second, and more importantly, the test query structure may not be visible during training. The generation process can help deal with this situation by learning the visible steps of the invisible complex query structures.

	\subsubsection{NN-based AQG Generator}
\begin{figure}
		\centerline{\includegraphics[width=0.5\textwidth]{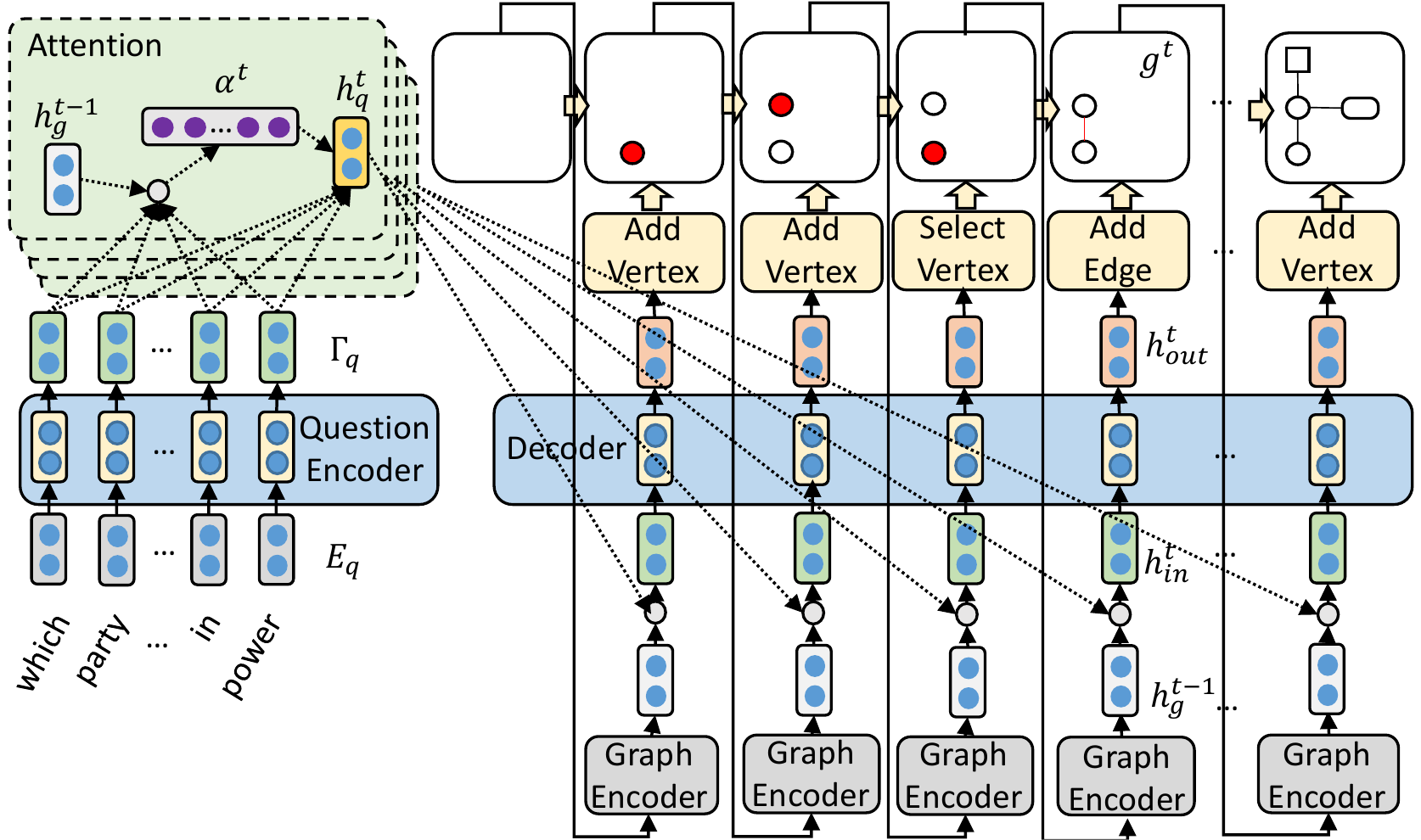}}
		\caption{ Architecture of the AQG generator. Red indicates the object being operated at each step.} \label{generation_model}
	\end{figure}

	The architecture of the proposed AQG generator is shown in Figure \ref{generation_model}. It takes a question and an empty graph $g^0$ as the inputs and then outputs a sequence of graphs $G = \{g^1, g^2, ..., g^L \}$, where  $g^L$ is the completed AQG. At each time step, the generator first predicts the argument for the operation and then generates a new graph by performing the operation on the previous graph. Intuitively, for each prediction, the generator needs to integrate the information of both the question and the previous graph. Therefore, the model consists of four components: a question encoder, a graph encoder, a decoder, and an executor.
	
	\paragraph{Question Encoder.} To capture the semantic information of the question, we apply a Bi-LSTM as the question encoder to get the vector representations of the question. First, we replace all entity (number)
	mentions (from linking results) used in the question $q$ by tokens $\langle e \rangle$ ($\langle n \rangle$), to enhance the generalization ability. Then, the question encoder works on the word embeddings of $q$ and outputs its vector representations $\Gamma_q = [\gamma^1_q,\gamma^2_q,...,\gamma^l_q]$, where $l$ denotes the number of the words in $q$.
	
	\paragraph{Graph Encoder.} To capture the semantic information of the graph generated at the previous time step, we leverage a graph transformer as the graph encoder. At time step $t$, the graph encoder embeds the graph $g^{t-1}$ and outputs a list of vertex vectors $\Gamma_v^{t-1} = [\gamma^1_v,\gamma^2_v,...,\gamma^n_v]$, a list of edge vectors $\Gamma_e^{t-1} = [\gamma^1_e,\gamma^2_e,...,\gamma^m_e]$ and a vector $h^{t-1}_g \in \mathbf{R}^d$, where $n$ and $m$ respectively denote the numbers of the vertices and edges in $g^{t-1}$, and $h^{t-1}_g$ is the representation of $g^{t-1}$.
	
	\paragraph{Decoder.} We employ a Bi-LSTM as the decoder to translate the information from the question and the previous graph into a vector at each time step. This vector carries the semantic information of the predicted argument. Specifically, at time step $t$, $h^{t}_q \in \mathbf{R}^d$ is the context vector from the question $q$, which is computed by
	\begin{equation}
	h^t_q = \sum_{i=1}^n \alpha^{t}_i \gamma^i_q
	\label{thesystem}
	\end{equation}
	\begin{equation}
	\alpha^{t}_i = \frac{\exp(h^{t-1}_g \cdot W\gamma^i_q) }{\sum_{j=1}^l \exp(h^{t-1}_g \cdot W\gamma^j_q)}
	\label{thesystem}
	\end{equation}
	where $\cdot$ denotes the dot product operation, $W \in \mathbf{R}^{d \times d}$ is the trainable parameter matrix, $h^{t-1}_g$ is the output from the graph encoder for $g^{t-1}$, and $\alpha^{t}_i$ is the weight of attention from $g^{t-1}$ for the $i$-th word of the question. This attention mechanism is based on the following intuition: At time step $t$, the model should focus on some parts of $q$ which do not carry the redundant information of $g^{t-1}$. i.e., avoid repeatedly adding the triple that has been added. Thereafter, the input to the decoder, denoted by $h_{in}^t \in \mathbf{R}^d$, is calculated by a skip connection: $h^t_{in} = h^{t-1}_g + h^{t}_q$. Here, $+$ denotes the element-wise addition. Although $h^{t-1}_g$ is already involved in calculating $h^{t}_q$, the decoder still needs the direct information of $g^{t-1}$ when making decisions, such as \textit{selectVertex}. Therefore, we adopt the idea from Residual Networks \cite{DBLP:conf/cvpr/HeZRS16}, enhancing the propagation of $h^{t-1}_g$ by this skip connection. Finally, the decoder takes $h_{in}^t$ as the input and outputs the semantic vector of the predicted argument, denoted by $h_{out}^t \in \mathbf{R}^d$.
	
	\paragraph{Executor.} At each time step $t$, the executor first transforms the decoder's output $h_{out}^t$ into the argument $*a_t$. Then, it performs the operation $f(g^{t-1}, *a^t)$ to output a new graph $g^t$. Corresponding to the types of the operator $f$, the executor contains three modules. For each module, we design a pointer network~\cite{DBLP:conf/nips/VinyalsFJ15} for predicting the argument:
	\begin{itemize}
		\item \textbf{Add Vertex Module.} The argument $c_v$ denotes the label of the added vertex, which is determined by
		\begin{equation}
		c_v = \arg\max_{i \in C_v} p_{av}(i|q, g^{t-1})
		\end{equation}
		\begin{equation}
		p_{av}(i|q, g^{t-1})= \frac{ \exp(h^t_{out} \cdot \beta_{i})}{\Sigma_{j \in C_v} \exp(h^t_{out} \cdot \beta_{j})}
		\end{equation}
		where $\beta_{i} \in \mathbf{R}^{d}$ is the parameter vector of label $i$ and $C_v =$\{``Ent", ``Type", ``Num",``Var", ``End"\}. Comparing to the AQG vertex classes, $C_v$ has one more class label ``End", which is used to terminate the process. Specifically, if ``End" is selected at time step $t$, the generation process ends and the generator outputs $g^t$. 
		\item \textbf{Select Vertex Module.} $v_{slc}$ denotes the index of the selected vertex of $g^{t-1}$, which is determined by
		\begin{equation}
		v_{slc} = \arg\max_{i \in V^{t-1}} p_{sv}(i|q, g^{t-1})
		\end{equation}
		\begin{equation}
	    p_{sv}(i|q, g^{t-1})= \frac{ \exp(h^t_{out} \cdot \gamma_v^i)}{\Sigma_{j \in V^{t-1}} \exp(h^t_{out} \cdot \gamma_v^j)}
		\end{equation}
		where $V^{t-1}$ denotes the vertex set of $g^{t-1}$, and $\gamma_v^j \in \mathbf{R}^{d}$ is the vector representation of the vertex $j$ in $g^{t-1}$ from the graph encoder.
		\item \textbf{Add Edge Module.} $c_e$ denotes the label of the added edge, which is determined by
		\begin{equation}
		c_e = \arg\max_{i \in C_e} p_{ae}(i|q, g^{t-1})
		\end{equation}
		\begin{equation}
		p_{ae}(i|q, g^{t-1})= \frac{ \exp(h^t_{out} \cdot \rho_{i})}{\Sigma_{j \in C_e} \exp(h^t_{out} \cdot \rho_{j})}
		\label{thesystem}
		\end{equation}
		where $\rho_{i} \in \mathbf{R}^{d}$ is the parameter vector of label $i$ and $C_e =$\{``Rel", ``Ord", ``Cmp", ``Cnt", ``Isa"\}.
	\end{itemize}
	
	\paragraph{KB Constraint.} AQG aims to be grounded to the KB to generate query graphs. Therefore, whether the AQG can be grounded in the knowledge base is also an important indicator to evaluate its correctness. To prevent the excessive expansion of the AQG, we utilize the KB to constrain its generation. Concretely, whenever the AQG completes an iteration, we ground it against the KB to check if there is at least one query graph that matches the AQG. If it exists, the generation process will continue, otherwise, the triple added by the latest iteration will be removed and the current graph is returned as the result. In addition, we employ a strategy to enhance the recognition of the type constraint: if there is no ``Type" vertex in the generated AQG but the type linking result set of the question is not empty, we attach a type vertex to a variable vertex and obtain a new AQG. Then, the KB constraint is required to verify the legality of this new AQG.
	
	\subsection{Training}
	In the experiments, each training sample consists of a question $q$ and its gold query graph $g_q^+$. The AQG corresponding to $g_q^+$ is regarded as the gold AQG, denoted by $g^+$.
	
	To get the ground-truth for training the AQG generator, we traverse $g^+$ to restore its generation process. In fact, for the same AQG, different traversal strategies and starting vertices result in different ground-truth. By experimental comparison, we adopt depth-first traversal. In addition, in a query graph, there is at least one variable vertex representing the answer\footnote{Other variables are hidden variables, like ?v1 in Figure \ref{one_question}.}, denoted by ?v0. Therefore, we select the vertex in $g^+$, which corresponds to ?v0, as the starting vertex of the traversal. Specifically, the ground-truth $\pi$ is obtained by the following steps: Initially, $\pi$ is an empty sequence. At the beginning of the traversal, the starting vertex $v_s$ is first visited and its label is added to $\pi$. Thereafter, whenever a fresh vertex $v$ is visited from the vertex $u$ along the edge $e$, the class label of $v$, the index of $u$, and the class label of $e$ will be added to $\pi$ in turn. When all vertices in $g^+$ are visited, the last label ``End" is added to $\pi$ and the ground-truth $\pi = \{a^1, a^2, ..., a^L \}$ is obtained. Here, $L$ is the step number of the generation process, and $a^t$ is the ground true argument (last argument) of each operation. During training time, our model is optimized by maximizing the log-likelihood of the ground true argument sequences:
	\begin{equation}
        \sum_{q \in Q} {\sum_{t=1}^L p(a^t|q, g^{t-1})}
	\end{equation}
	Here, $Q$ is the set of the questions, and $p(a^t|q, g^{t-1})$ is the predicted probability of the ground true argument $a^t$, including $p_{av}$, $p_{sv}$ and $p_{ae}$.
	
	\section{Experiments}
	\subsection{Experimental Setup}
	Our models are trained and evaluated over the following three KBQA datasets:
	
	\textbf{LC-QuAD} \cite{DBLP:conf/semweb/TrivediMDL17} is a gold standard complex question answering dataset over the DBpedia 04-2016 release, having 5,000 NLQ and SPARQL pairs. The dataset is split into 4,000 training and 1,000 testing questions\footnote{\url{https://figshare.com/projects/LC-QuAD/21812}}.
	
	\textbf{ComplexQuestions} (CompQ) \cite{DBLP:conf/coling/BaoDYZZ16} is a question answering dataset over Freebase, which contains 2,100 complex questions collected from Bing search query log. The dataset is split into 1,300 training and 800 testing questions\footnote{\url{https://github.com/JunweiBao/MulCQA/tree/ComplexQuestions}}.
	
	\textbf{WebQuestions} (WebQ) \cite{DBLP:conf/emnlp/BerantCFL13} is a question answering dataset over Freebase, which contains 5,810 questions collected from Google Suggest API. It is split into 3,778 training and 2,032 testing QA pairs\footnote{\url{https://nlp.stanford.edu/software/sempre/}}. Since more than 80\% of the questions in this dataset are simple questions that contain only one fact, we use it as a benchmark for simple questions.
	
	For LC-QuAD, we directly use its provided standard SPARQL as the gold query. However, ComplexQuestions and WebQuestions only contain question-answer pairs, so we search the candidate queries released by \cite{DBLP:conf/emnlp/LuoLLZ18} and take the query with the highest F1-score as the gold query. To compare with previous methods\cite{DBLP:conf/emnlp/LuoLLZ18,DBLP:conf/semweb/MaheshwariTLCF019}, we follow them to use the results of the state-of-the-art entity-linking tool S-MART \cite{DBLP:conf/acl/YangC15} on CompQ and WebQ, and use the gold entities on LC-QuAD. For each dataset, we randomly select the $10$\% of the training set as the development set.
	
	\paragraph{Implementation details.} In our experiments, all word vectors are initialized
	with 300-d pretrained word embeddings using GloVe\cite{DBLP:conf/emnlp/PenningtonSM14}. The following hyper-parameters are tuned on development sets: (1) Both the sizes of the hidden states are set to 256; (2) The layer number of Bi-LSTM is set to 1 and the layer number of graph Transformer is set to 3; (3) The learning rate is set to $2 \times 10^{-4}$; (4) The number of the training epochs is set to $30$. Our code are publicly available\footnote{\url{https://github.com/Bahuia/AQGNet}}.
	
	\subsection{End-to-End Results}
	We compared our approach with several existing KBQA methods. \cite{DBLP:conf/www/AbujabalYRW17} performs query building by existing templates. \cite{DBLP:conf/acl/YihCHG15,DBLP:conf/coling/BaoDYZZ16,DBLP:conf/acl/YuYHSXZ17} construct pipelines to generate query graphs. \cite{DBLP:conf/emnlp/LuoLLZ18,DBLP:conf/semweb/MaheshwariTLCF019} obtain state-of-the-art performance on CompQ and LC-QuAD by query ranking, respectively. \cite{DBLP:conf/emnlp/Hu0Z18} achieves a state-of-the-art result on WebQ by state-transition over dependency parsing. The main difference between our method and theirs is that they perform state transition by enumerating pre-defined transition conditions, but we let our model learn to how to expand the AQG automatically at each step.
		\begin{table} \label{end_to_end_results}
		\begin{center}
			\scalebox{0.9}{
			\begin{tabular}{cccc}
				\hline
				\rule{0pt}{12pt}
				Method & LC-QuAD & CompQ & WebQ
				\\
				\hline
				\\[-6pt]
				\cite{DBLP:conf/acl/YihCHG15} & - & 36.9 & 52.5 \\
				\cite{DBLP:conf/coling/BaoDYZZ16} & - & 40.9 & 52.4 \\
				\cite{DBLP:conf/www/AbujabalYRW17} & - & - & 51.0 \\
				\cite{DBLP:journals/pvldb/CuiXWSHW17} & - & - & 34.0 \\
				\cite{DBLP:conf/emnlp/LuoLLZ18} & - & 42.8 & 52.7 \\
				\cite{DBLP:conf/emnlp/Hu0Z18} & - & - & \textbf{53.6} \\
				\cite{DBLP:conf/acl/YuYHSXZ17} & 70.0 & - & - \\
				\cite{DBLP:conf/semweb/MaheshwariTLCF019} & 71.0 & - & - \\
				\hline
				\\[-6pt]
				\quad Our approach & \textbf{74.8} & \textbf{43.1} & 53.4 \\
				\hline
				\\[-6pt]
			\end{tabular}}
			{\caption{Average F1-scores on LC-QuAD, CompQ and WebQ.}\label{end_to_end_results}}
		\end{center}
	\end{table}
	\begin{table*}
		\begin{center}
			\scalebox{0.9}{
				\begin{tabular}{lcccccccccccc}
					\hline
					\rule{0pt}{12pt}
					
					&\multicolumn{4}{c}{LC-QuAD}&\multicolumn{4}{c}{CompQ}&\multicolumn{4}{c}{WebQ}\\
					\cline{2-13}
					\\[-6pt]
					&$N_c$ &P &R &F1 &$N_c$ &P &R &F1 &$N_c$ &P &R &F1 \\
					\hline
					\\[-6pt]
					\quad our approach &\textbf{204.4} &\textbf{75.63} &75.01 &\textbf{74.75} &\textbf{88.3} &\textbf{42.11} &54.59 &\textbf{43.06} &\textbf{76.1} &\textbf{53.77} &\textbf{61.81} &\textbf{53.43} \\
					\hline
					\quad replacing AQG with ST
					&1379.9 &65.89 &\textbf{75.30} &69.53 
					&204.4 &40.76 &\textbf{54.98} &42.47 
					&147.5 &52.16 &61.25 &52.73 \\
					\hline
					\\[-6pt]
			\end{tabular}}
			{\caption{Average candidate number($N_c$), precision (P), recall (R) and F1-scores (F1) for different candidate query generation methods.}\label{table2}}
		\end{center}
	\end{table*}
	
	The experimental results are reported in Table \ref{end_to_end_results}. Our approach achieves state-of-the-art results on both LC-QuAD and CompQ, and ranks second only to \cite{DBLP:conf/emnlp/Hu0Z18} on WebQ. Although the performance on WebQ did not exceed \cite{DBLP:conf/emnlp/Hu0Z18}, our approach achieves a result that closes to theirs on WebQ without any predefined transition conditions. From another perspective, our approach outperforms all the existing ranking-based approaches \cite{DBLP:conf/emnlp/LuoLLZ18,DBLP:conf/semweb/MaheshwariTLCF019} on the three datasets.
	
	The performance of \cite{DBLP:conf/www/AbujabalYRW17,DBLP:conf/coling/BaoDYZZ16,DBLP:conf/acl/YuYHSXZ17} is limited by the error propagation of the pipeline for ranking components. \cite{DBLP:conf/emnlp/LuoLLZ18,DBLP:conf/semweb/MaheshwariTLCF019} rank entire queries to joint the information of components, thus achieve better results. However, enumeration on query structures makes their performance suffer from noisy queries. Existing models lack sufficient representation capacity to balance structure and components. Our approach decouples the query ranking task and exploits a specific generative model to handle the information of structures, achieving better results.
	
	\subsection{Detailed Analysis}
	
	\subsubsection{Contribution of AQG Constraint}
	In order to evaluate the improvement by AQG constraint, we kept our baseline ranking model unchanged but generated candidate queries by the staged-transition strategy (ST).
	
	Table \ref{table2} shows the average numbers of candidate queries, precision, recall, and F1-score of the entire systems based on different candidate generation strategies. Intuitively, the size of the candidate set generated by AQG is much smaller than that generated by ST, especially on LC-QuAD. In addition, comparing to the ST-based strategy, the AQG-based strategy improves the performance of the entire system on all the datasets by increasing the average precision of answering.
	
	To further analyze the impact of question complexity on the performance, we divided the questions in each dataset into four complexity levels according to the number of edges in the correct query graph. Here, level $i$ indicates that the query graph contains $i$ edges. We tested the average F1-scores when answering the questions of different levels. Figure \ref{figure7}(a),\ref{figure7}(b) and \ref{figure7}(c) show the results. The improvements of the AQG-based strategy over the ST-based strategy increase with the complexity of questions. Furthermore, we found that the performance of both methods on CombQ and WebQ (using entity linking results) increases as the question is more complex, but it is not reflected in LC-QuAD (using gold entities). It hints to the fact that constraints in complex questions can provide more comprehensive information for entity linking.
	
	\subsubsection{Ablation Study on AQG Generator}
	
	To explore the contributions of various
	components of our AQG generator, we compared the following settings:
	
	\paragraph{w/o attention.} We replaced the attention mechanism with an average-pooling to get $h^{t}_q$.
	
	\paragraph{w/o skip connection.} We removed the skip connection of $h^{t-1}_g$ and $h^{t}_q$ and only used $h^{t}_q$ as the input to the decoder.
	
	\paragraph{w/o graph encoder.} We removed the graph encoder, that is, at decoding time step $t$, we use $h_{out}^{t-1}$ to calculate $h^{t}_q$ and obtain $h_{in}^t$ by the element-wise addition of $h_{out}^{t-1}$ and $h^{t}_q$.
	
	\paragraph{w/o KB constraint.} We removed the KB constraint that prevents the excessive expansion of AQG.
	\begin{table}
		\begin{center}
			\scalebox{0.9}{
			\begin{tabular}{lccc}
				\hline
				\rule{0pt}{12pt}
				&\multicolumn{1}{c}{LC-QuAD}&\multicolumn{1}{c}{CompQ}&\multicolumn{1}{c}{WebQ}\\
				\hline
				\\[-6pt]
				AQG generator &\textbf{83.00} &\textbf{88.88} &91.88 \\
				\hline
				\\[-6pt]
				w/o attention     &80.20 &87.75 &90.50 \\
				w/o skip connection &78.40 &85.45 &91.29 \\
				w/o graph encoder &38.60 &39.13 &89.90 \\
				w/o KB constraint &72.90 &80.52 &90.36 \\
				\hline
				\\[-6pt]
				breadth-first   &81.80 &84.12 &\textbf{91.95} \\
				random          &79.80 &83.78 &91.55 \\
				\hline
				\\[-6pt]
			\end{tabular}}
			{\caption{Accuracy of AQG generation.}\label{table3}}
		\end{center}
	\end{table}
	
	The middle of Table \ref{table3} shows the accuracy of AQG generation of different settings. By removing the attention mechanism, the accuracy declined approximately 2\% on the three datasets. Removing the graph encoder results in a significant drop in performance, especially on LC-QuAD and CombQ. It demonstrates that the structural information of the previous graph is essential. The performance drop by removing skip connection also illustrates this fact. The main reason for the smaller drop on WebQ can be that most questions in WebQ are simple questions whose true query only contains one edge. In this case, the model can also achieve good results only by the memory of the previous operation. Rmoving the KB constraint makes the performance approximately drop 10\% over both the complex datasets, which reveals that KB is an effective tool for the query structure disambiguation.
	
	\subsubsection{Impact of the Ground-truth Construction Strategy}
	We also evaluate the accuracy of the AQG generator supervised by different ground-truth, which are constructed by following traversal strategies.
	\begin{figure*}
		\centerline{\includegraphics[width=\textwidth]{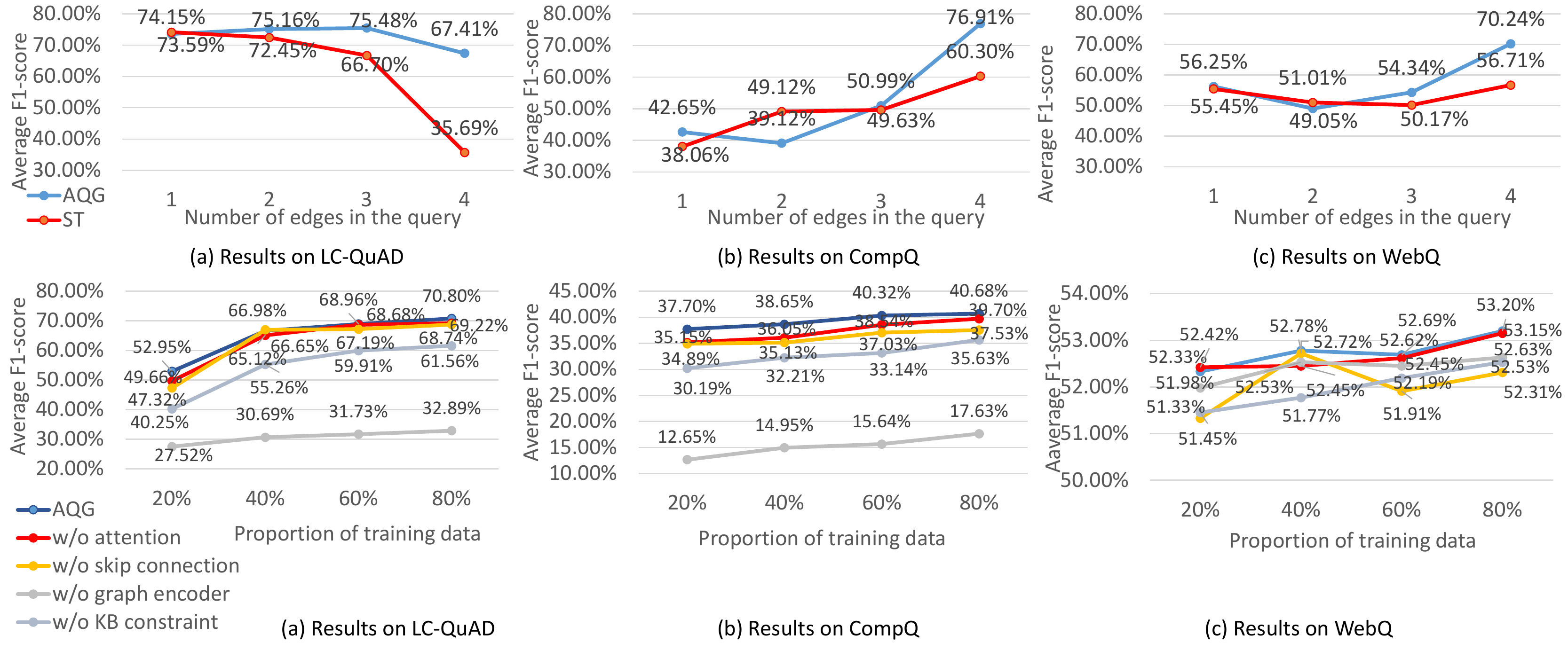}}
		\caption{F1-score with varied complexity levels of questions and proportions of training data} \label{figure7}
	\end{figure*}
	
	\paragraph{Breadth-first traversal.} We replaced the depth-first traversal with the breadth-first traversal.
	
	\paragraph{Random traversal.} We adopted the random traversal. Specifically, once visiting a vertex $v$, randomly select an unvisited vertex connected to $v$ as the next vertex to visit.
	
	The bottom of Table \ref{table3} shows the results. The performance is worst when the ground-truth is constructed by random traversal. Depth-first traversal achieves better performance than breadth-first traversal on LC-QuAD and CompQ. By observation, we speculate it is because the facts of complex questions are typically reflected in the form of chains. Depth-first traversal can effectively preserve this chain structure so that it achieved better results. Three traversal strategies achieve close results on WebQ because three traversal strategies have the same results for simple query graphs.
	
	\subsubsection{Results on Varied Sizes of Training Data}
	We tested the performance of our approach with different sizes of training data. The results are shown in Figure \ref{figure7}(d),\ref{figure7}(e) and \ref{figure7}(f). The model equipped with the attention mechanism, the graph encoder, the skip connection, and the KB constraint always maintains optimal performance with different sizes of training data. Furthermore, with less training data, the performance of our approach can still maintain certain stability. It is due to the fact that our model learns fine-grained generative steps instead of entire structures. In this way, even if there are some invisible test query structures, which do not appear in the training data, our model can still handle them by visible generative steps.
	
	\section{Related Work}
	Traditional semantic parsing based KBQA approaches \cite{DBLP:conf/acl/BerantL14,DBLP:conf/kdd/FaderZE14,DBLP:conf/emnlp/ReddyTPSL17} aim to learn semantic parsers that translate natural language questions into a logical form expression. Recent researches proposed an alternate approach to semantic parsing by treating KBQA as a problem of semantic graph generation. \cite{DBLP:conf/acl/YihCHG15} defines a sequence of stages for generating query graphs, each relying on a set of manually defined rules for how query conditions are added. \cite{DBLP:conf/acl/YuYHSXZ17} proposes a hierarchical representation model for the matching of question and KB relations. Current KBQA methods simplify the pipeline into query ranking. \cite{DBLP:conf/emnlp/LuoLLZ18} apply a staged candidate generation strategy to generate candidate query graphs and then rank them by utilizing a semantic matching model.  \cite{DBLP:conf/semweb/MaheshwariTLCF019} follows the candidate generation methods proposed in \cite{DBLP:conf/acl/YihCHG15} and propose a novel self-attention based slot matching model. These approaches try to capture the structural information and the semantic information by a single model. The main difference between our approach and previous approaches is that we explicitly leverage the structural constraint (AQG) to narrow the search space of candidate queries, in order to release the burden of query ranking. In addition, to avoid using any predefined structured templates which limit the generalization ability of our approach, we propose a generative framework to generate AQG based on basic graph-level operations. To the best of our knowledge, this is the first attempt to predict the query structures without any pre-defined templates. Last but not least, to utilize the structural information of the graph to help the model make decision in the AQG generation process, we apply graph transformer\cite{DBLP:conf/naacl/Koncel-Kedziorski19}, which is an existing graph neural network model, to learn the representation of the AQG.
	The most related work to ours is \cite{DBLP:conf/emnlp/Hu0Z18} and \cite{DBLP:conf/emnlp/DingHXQ19}. \cite{DBLP:conf/emnlp/Hu0Z18} proposes a semantic query graph to constrain candidate query generation. They perform state transition on the dependency tree by predefined conditions. In contrast, we make the model learn to perform the state transition without any conditions. \cite{DBLP:conf/emnlp/DingHXQ19} proposes to leverage the query substructure templates in the training data to construct complex queries.  Different from \cite{DBLP:conf/emnlp/DingHXQ19}, we try to predict the query structure from the perspective of graph generation without using templates.
	
	\section{Conclusion}
	 In this paper, we presented our two-stage formal query building approach. Our approach can automatically predict the query structure and use it as a constraint to avoid generating noisy candidate queries, thereby improving the performance of the following query ranking. The experimental results showed that our approach achieved superior results than the existing methods on complex questions, and produced competitive results on other simple question based datasets.
	 In future work, We will try to extend the AQG generation framework in order to deal with the query of general graph structures. In addition, we plan to apply reinforcement learning instead of supervised learning to train the AQG generator.
	 
	 \section*{Acknowledgments}
	 The work is partially supported by the Natural Science Foundation of China under grant No. 61502095, National Key Research and Development Program of China under grants (2018YFC0830200, 2017YFB1002801), the Natural Science Foundation of China grants (U1736204, 61602259), the Judicial Big Data Research Centre, School of Law at Southeast University and the project no. 31511120201.

\bibliographystyle{named}
\bibliography{ijcai20}

\end{document}